\documentclass[letterpaper]{article} 
\usepackage{aaai24}  
\usepackage{times}  
\usepackage{helvet}  
\usepackage{courier}  
\usepackage[hyphens]{url}  
\usepackage{graphicx} 
\usepackage{natbib}  
\usepackage{caption} 
\usepackage{algorithm}
\usepackage{algorithmic}
\usepackage{bm}
\usepackage{amsmath}
\usepackage{amssymb}
\usepackage{booktabs}
\usepackage{newfloat}
\usepackage{listings}
\usepackage{color}
\usepackage{booktabs}
\usepackage{multirow}
\usepackage{subfig}

\DeclareCaptionStyle{ruled}{labelfont=normalfont,labelsep=colon,strut=off} 
\floatstyle{ruled}
\newfloat{listing}{tb}{lst}{}
\floatname{listing}{Listing}
\title{AdapEdit: Spatio-Temporal Guided Adaptive Editing Algorithm for Text-Based Continuity-Sensitive Image Editing}
\author{
    Zhiyuan Ma\textsuperscript{\rm 1,2}\thanks{Equal contributions.}, Guoli Jia\textsuperscript{\rm 3}\footnotemark[1], Bowen Zhou\textsuperscript{\rm 1}\thanks{Corresponding author.}
}
\affiliations{
    \textsuperscript{\rm 1}Department of Electronic Engineering, Tsinghua University\\

    \textsuperscript{\rm 2}Shanghai Artificial Intelligence Laboratory\\

    \textsuperscript{\rm 3}College of Computer Science, Nankai University

    \{mzyth, zhoubowen\}@tsinghua.edu.cn, exped1230@gmail.com
}

\begin{document}
\maketitle

\begin{abstract}
With the great success of text-conditioned diffusion models in creative text-to-image generation, various text-driven image editing approaches have attracted the attention of many researchers. However, previous works mainly focus on discreteness-sensitive instructions such as adding, removing or replacing specific objects, background elements or global styles (\emph{i.e., ``hard editing''}), while generally ignoring subject-binding but semantically fine-changing continuity-sensitive instructions such as actions, poses or adjectives, and so on (\emph{i.e., ``soft editing''}), which hampers generative AI from generating user-customized visual contents. To mitigate this predicament, we propose a spatio-temporal guided \underline{\textbf{ada}}ptive \underline{\textbf{edit}}ing algorithm AdapEdit, which realizes adaptive image editing by introducing a soft-attention strategy to dynamically vary the guiding degree from the editing conditions to visual pixels from both temporal and spatial perspectives. Note our approach has a significant advantage in preserving model priors and does not require model training, ﬁne-tuning, extra data, or optimization. We present our results over a wide variety of raw images and editing instructions, demonstrating competitive performance and showing it significantly outperforms the previous approaches. Code is available: https://github.com/AnonymousPony/adap-edit.
\end{abstract}

\section{Introduction}

Text-conditioned image synthesis has recently received tremendous success with various magnificent and creative text-to-image cases being produced, where diffusion probability models (DPMs)~\cite{ho2020denoising,nichol2021improved}, as the backbone network that plays an important role, has been developed into a variety of text-to-image generation applications~\cite{saharia2022photorealistic,ramesh2022hierarchical,ma2023follow}. However, the controllability of image synthesis and the support for user-customization are still challenging issues and have received more and more researchers' attention. To solve these problems, Stable Diffusion~\cite{rombach2022high} and ControlNet~\cite{zhang2023adding}, as two leading studies, have proposed two efficient text-conditioned learning frameworks for training, and subsequently, a series of text-guided image editing methods~\cite{hertz2022prompt,ruiz2023dreambooth,wallace2023edict,kawar2023imagic,mokady2023null} have been proposed, which achieve user-customized image output by editing or modifying the original image given the human-written instructions, and have witnessed a huge improvement in terms of controllability.

\begin{figure}[t]
	\centering
	\includegraphics[width=1\linewidth]{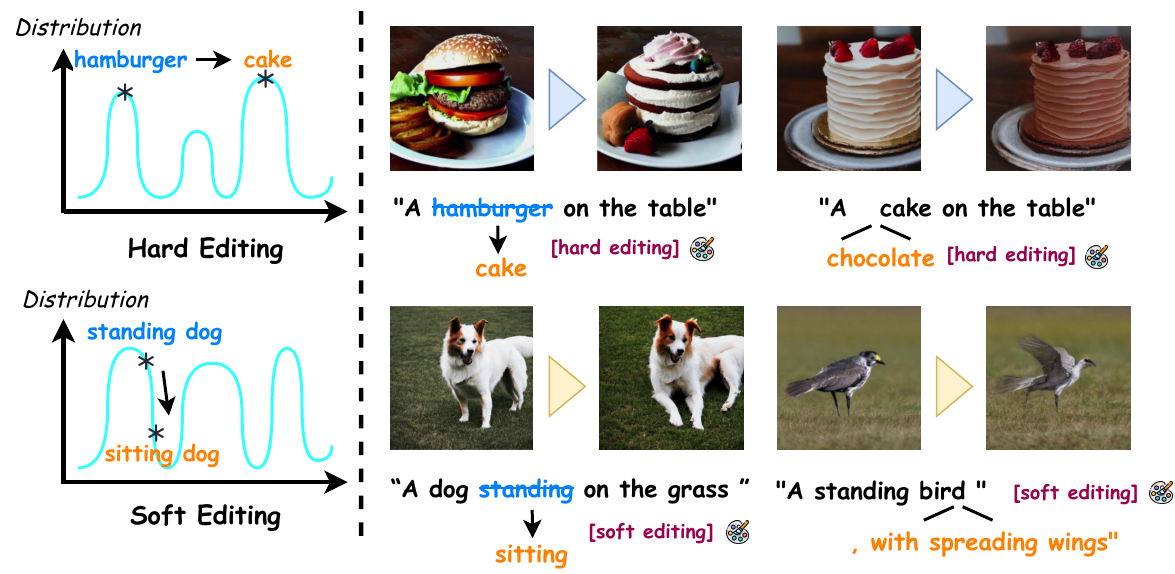}
	\caption{Example of image editing to show discreteness-sensitive image manipulations (\emph{i.e., hard editing}) and continuity-sensitive image manipulations (\emph{i.e., soft editing}).}
	\label{fig: intro}
\end{figure}

\begin{figure*}[ht!]
	\centering
	\includegraphics[width=1\textwidth]{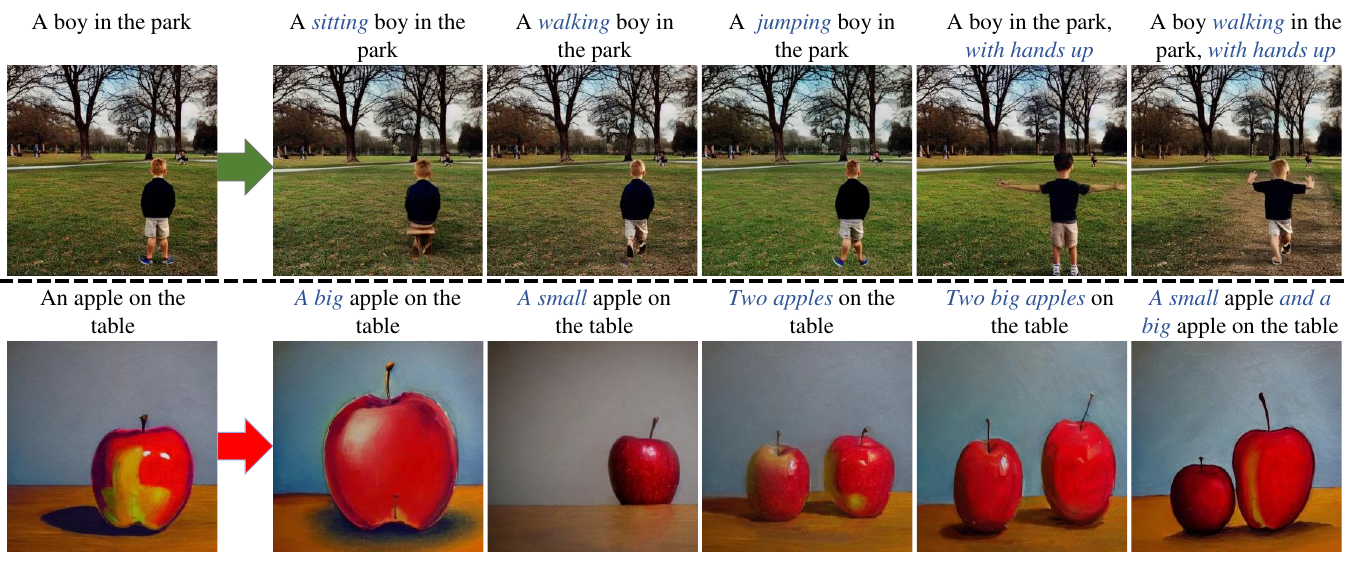}
	\caption{The performance of AdapEdit with soft editing instructions.
    The leftmost images are directly generated by the original condition, images on other lines are 
    edited by the original and editing conditions.}
	\label{fig: motivation}
\end{figure*}

Though achieving remarkable progress, existing text-based image editing methods still suffer from the following three limitations. (1) \emph{Soft editing dilemma.}~\textbf{Firstly}, as illustrated in Figure~\ref{fig: intro}, we notice that the previous approaches mainly focus on relatively simple editing instructions such as adding, removing or replacing specific objects, background elements or global styles (\emph{i.e., hard editing}), while generally ignoring subject-binding but semantically fine-changing complex instructions such as actions, poses or adjectives, and so on (\emph{i.e., soft editing}), which hampers generative AI from generating user-customized images. (2) \emph{Spatio-temporal continuity dilemma.}~\textbf{Secondly}, we notice that the previous methods generally adopt the same guide degree during the text-to-image diffusion process for each pixel of the original image (\emph{i.e., guided target}) and each word of the textual condition (\emph{i.e., guiding source}), which may result in artifacts emerged in the unexpected areas, as well as may also cause semantic discontinuities with surrounding pixels to appear around the edited objects. (3) \emph{Training dilemma.}~\textbf{Thirdly}, to improve model's controllable generation ability under given editing-instructions, most of the previous methods need to feed a large number of labeled image-text pairs as training data, and also need to design additional optimization objectives for training. However, the costly training often causes the \emph{language drift} phenomenon appears in image priors conditioned on the editing instructions training, whereas forget the priors under the original unedited instructions.

To address the aforementioned limitations, we propose \emph{AdapEdit}, a spatio-temporal guided adaptive editing algorithm for complex continuity-sensitive image editing tasks. Specifically, to address the first limitation, we introduce a soft attention strategy into algorithm to adaptively guide the attention degrees from textual words to visual pixels for controllable text-to-image editing. Based on the soft attention strategy, we further address the second limitation of \emph{continuity dilemma} by respectively designing a flexible word-level temporal (FWT) adjustment module and a dynamic pixel-level spatial (DPS) weighting module to achieve spatio-temporal guided adaptive editing through spatial guiding scales $\textbf{s}_\texttt{[V]}$ and temporal guiding scales $\tau_{c^*}$. Moreover, we introduce a spatial interpolation weight $\lambda_{S}$, which is a hyper-parameter and is adopted to finely control the editing amplitude at the pixel level. Finally, the entire AdapEdit algorithm is executed in only one forward diffusion process and does not require any training or optimization with the algorithm time-overhead is as low as $O(n)$, which naturally solves the third \emph{training dilemma}.
Our main contributions are summarized as follows:
\begin{itemize}
    \item \textbf{Soft editing capabilities enhancement.} The ability to support more complex editing tasks, such as soft editing tasks, has been enhanced for better user-customized image generation scenarios.
    \item \textbf{Editing naturalness and contextual continuity improvement.} By assigning variable spatio-temporal guidance scales to adaptively guide the attentions from textual tokens to visual pixels, the naturalness and contextual semantic coherence of the edited images are significantly improved.
    \item \textbf{No training required and no damage to model priors.} Since a deterministic adaptive algorithm is developed, it requires almost no training costs such as training data, annotations, additional optimization targets, and huge GPU overheads. Moreover, more importantly, our algorithm does not break the priors of the DPMs themselves.
\end{itemize}

\begin{figure*}[t]
	\centering
	\includegraphics[width=1.0\linewidth]{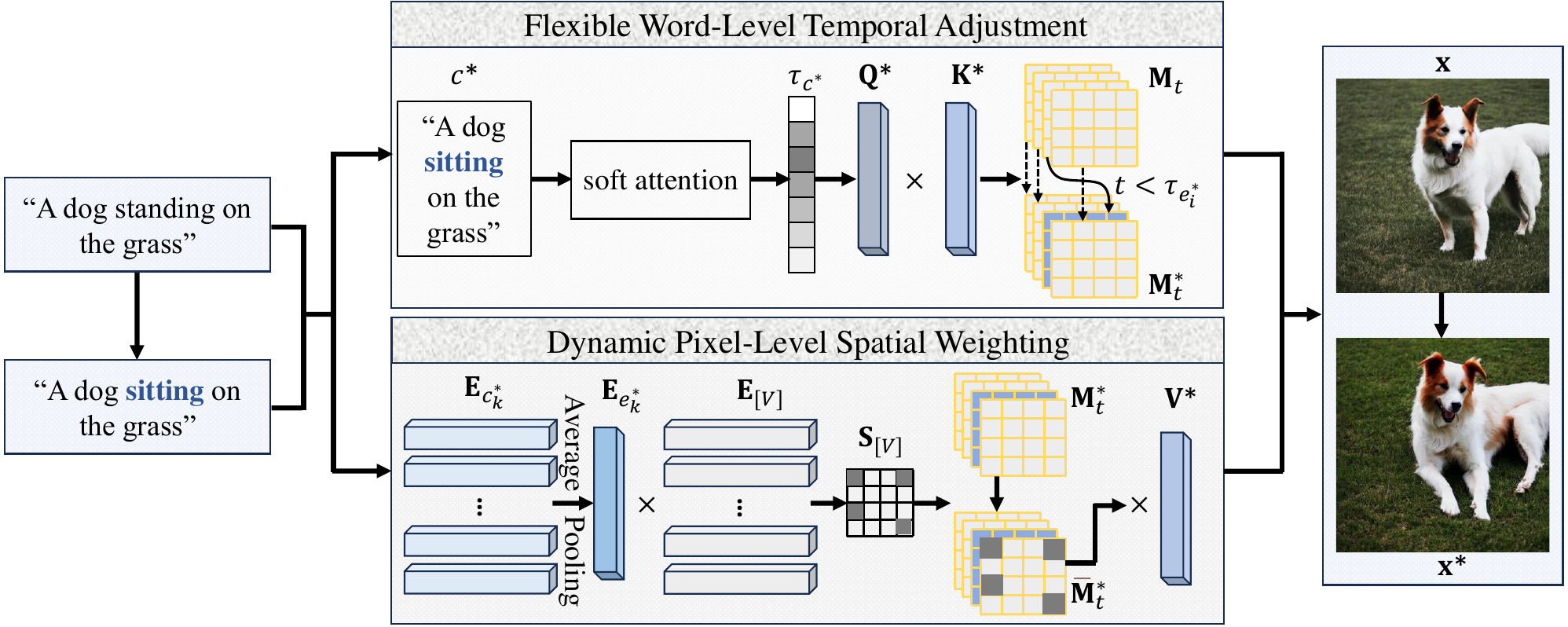}
	\caption{The framework overview of the proposed AdapEdit algorithm.}
	\label{fig: model}
\end{figure*}

\section{Related Work}

\noindent\textbf{Diffusion-Based Generative Models.} Recent years has witnessed the remarkable success of diffusion-based generative models, due to their excellent performance in the diversity and impressive generative capabilities. These previous efforts mainly focus in sampling procedure~\cite{song2020score,liu2022pseudo}, conditional guidance~\cite{dhariwal2021diffusion,nichol2021glide}, likelihood maximization~\cite{kingma2021variational,kim2022maximum} and generalization ability~\cite{kawar2022denoising,ma2023lmd} and have enabled state-of-the-art image synthesis. \\

\noindent\textbf{Text-Guided Image Synthesis Models.}
Text-guided image synthesis aims to generate images conforming to the text semantics by sampling and predicting the noise distributions in diffusion models. Take advantage of guidance diffusion techniques, recent large text-to-image models such as Imagen~\cite{saharia2022photorealistic}, DALL-E2~\cite{ramesh2022hierarchical}, Parti~\cite{yu2022scaling}, CogView2~\cite{ding2022cogview2} and Stable Diffusion~\cite{rombach2022high}, have synthesized a wide variety of unprecedented images and shown excellent performance. However, it is worth noting that these models only support sampling in a random Gaussian distribution conditioned on given text prompts, whereas cannot achieve customized image generation through directly editing on the user-uploaded images, which hampers from supporting more complex and challenging image customizations tasks.\\

\noindent\textbf{Text-Guided Image Editing Models.}
To solve the problem of controllability and support customizations, a series of diffusion-based image editing models have been proposed. As a leading research in controllable text-to-image generation, \textbf{ControlNet}~\cite{zhang2023adding} proposes an end-to-end neural architecture that controls aforementioned large diffusion models (e.g., Stable Diffusion) to learn task-specific input conditions for efficiently supporting condition-guided image synthesis.
Based on this effective fine-tuning framework, a series of image editing methods have been proposed and have advanced the controllability research. \textbf{Prompt-to-Prompt}~\cite{hertz2022prompt} is one of the most representative models, which implements directed editing by performing a series of hard operations such as insert, replace, and reweight over attention maps of the cross-attention layer. While it still suffers from the accumulated errors caused by classifier-free guidance. To address the issue, \textbf{Null-Text-Inversion}~\cite{mokady2023null} introduces pivotal inversion technique to rectify the accumulated errors and proposes a null text optimization scheme for better performance. Subsequently, to implement subject-binding editing, \textbf{DreamBooth}~\cite{ruiz2023dreambooth}~proposes to introduce a unique identifier for subject-specific binding, which is fine-tuned on a pretrained diffusion model with a class-specific prior preservation objective.
Meanwhile, \textbf{Imagic} also utilizes a pre-trained text-to-image diffusion model and first uses it to ﬁnd a optimal text embedding for the input image by ﬁne-tuning the diffusion model, then adopts linearly interpolate technique to obtain a semantically meaningful mixture for effective editing.

Despite remarkable success, these recent image editing efforts still suffer from the aforementioned dilemmas, \emph{soft editing}, \emph{spatio-temporal continuity}, and \emph{training
dilemma}, which prevent generative models from moving toward more fine-grained control and more universal customization. To the end, we propose AdapEdit algorithm, which can be applied to all of the above models to improve their editing performance and can be applied to cope with more complex editing instructions (i.e., continuity-sensitive instructions) such as postures, actions, adjectives, and other subject-binding but semantically fine-changing instructions. 
\\

\section{Methodology}
\subsection{Preliminaries}
\paragraph{Diffusion models.} The diffusion model is modeled as: 1) a deterministic forward noising process $q(\textbf{x}_{t}|\textbf{x}_{t-1})=\mathcal{N}(\textbf{x}_t;\sqrt{\alpha_t}\textbf{x}_{t-1},(1-\alpha_t)\textbf{I})$ from the original image $\textbf{x}_0$ to a pure-Gaussian distribution $\textbf{x}_T\sim\mathcal{N}(0,\textbf{I})$, which can be formulated in an accumulated form as follows:
\begin{equation}
\textbf{x}_t=\sqrt{\bar{\alpha}_t}\textbf{x}_0+\sqrt{1-\bar{\alpha}_t}\bm{\epsilon},\quad \bm{\epsilon}\sim\mathcal{N}(0,\textbf{I})
\end{equation}
2) and a iteratively predictable reverse denoising process $p_\theta(\textbf{x}_{t-1}|\textbf{x}_t)=\mathcal{N}(\textbf{x}_{t-1};\bm{\mu}_{\theta}(\textbf{x}_{t},t),\bm{\Sigma}_{\theta}(\textbf{x}_{t},t))$, which can be trained in a simplied denoising objective $\mathcal{L}_{\text{simple}}$ by merging $\bm{\mu}_{\theta}$ and $\bm{\Sigma}_{\theta}$ into predicting noise $\bm{\epsilon}_\theta$,
\begin{equation}
\mathcal{L}_{\text{simple}}=E_{\textbf{x}_0,t,\bm{\epsilon}\sim\mathcal{N}(0,\textbf{I})}[||\bm{\epsilon}-\bm{\epsilon}_\theta(\textbf{x}_t,t)||^2_2]
\end{equation}
where $t\sim\mathcal{U}[1,T]$ is time parameters, $\mathcal{U}(\cdot)$ denotes uniform distribution. Moreover, in Stable Diffusion~\cite{rombach2022high}, the image $\textbf{x}_{t}$ is compressed into a latent variable $\textbf{z}_{t}$ by encoder $\mathcal{E}$ for more efficient training, i.e., $\textbf{z}_{t}=\mathcal{E}(\textbf{x}_{t})$, thus this preliminary objective is usually defined as making $\bm{\epsilon}_\theta(\textbf{z}_t,t)$ as close to $\bm\epsilon\sim\mathcal{N}(0,\textbf{I})$ as possible.

\paragraph{Text-guided diffusion models.} The core of text-guided diffusion models is to integrate the semantics of text condition $\bm c$ into noise prediction model $\bm{\epsilon}_\theta(\textbf{z}_t,t)$ to generate visual contents conforming to text semantics, i.e., $\bm{\epsilon}_\theta(\textbf{z}_t,t,\bm c)$. The classifier-free guidance technique has recently been widely adopted in text-guided image generation as,
\begin{equation}
    \tilde{\bm{\epsilon}}_\theta(\textbf{z}_t,t,\bm c,\varnothing)=w\cdot \bm{\epsilon}_\theta(\textbf{z}_t,t,\bm c)+(1-w)\cdot \bm{\epsilon}_\theta(\textbf{z}_t,t,\bm{\varnothing})
\end{equation}
where $w=7.5$ is default linear parameter for weighting the unconditional guidance objective and conditional guidance objective in Stable Diffusion, $t$ is time input, $\bm{c}$ is text condition, $\bm\varnothing$ denotes null text embedding initialized by zero vector and $\theta$ is model parameters. Note that all of these parameters will be individually or jointly optimized for controlled image editing in diffusion-based variants. 

\begin{algorithm}[tb]
\caption{AdapEdit algorithm}
\label{algorithm}
\textbf{Input}: The text condition $\bm{c}$, editing instruction $\bm{c}^*$\\
\textbf{Output}: Image $\textbf{x}$, edited image $\textbf{x}^*$
\begin{algorithmic}[1] 
\STATE \textbf{Initialize}: Random seed and parameters
\STATE \textbf{Function FWT:}
\IF {last diffusion step}
\STATE Calculate the average of the cross-attention maps $\textbf{M}^{{\bm{c}^*}}_t$;
\STATE Multiply($\textbf{M}^{{\bm{c}^*}}_t$, $\textbf{E}_{v}$) $\rightarrow$ $\textbf{E}_{\bm{c}^*}$ (Eq.5);
\STATE Average Pooling($\textbf{E}_{\bm{e}^*_i}$, $i \in \bm{c}^*_k$) $\rightarrow$ $\textbf{E}_{\bm{e}^*_k}$;
\STATE Multiply($\textbf{E}_{\bm{c}^*}$, $\textbf{E}_{\bm{e}^*_k}$) $\rightarrow$ $\bm{A}_{\bm{c}^*}$ (Eq.6);
\FOR{$i \in N_{\bm{c}^*}$}
\STATE Update $\tau_{\bm{e}^*_i}$ (Eq.7);
\ENDFOR
\ENDIF
\STATE \textbf{Function DPS:}
\FOR{t = $T$, $T-1$,...,1}
\STATE Multiply($\textbf{E}_{\bm{e}^*_k}$, $E_v$) $\rightarrow$ $\textbf{S}_\texttt{[V]}$ (Eq.9);
\STATE Update $\bm{\mathcal{C}}$ ($\textbf{M}_t^{\bm{c}}$, $\textbf{M}_t^{\bm{c}^*}$) (Eq.8);
\STATE Update($\overline{\textbf{M}}_t^{\bm{c}^*}$) (Eq.4);
\ENDFOR
\STATE Generate \{$\textbf{x}$, $\textbf{x}^*$\};
\STATE \textbf{return} \{$\textbf{x}$, $\textbf{x}^*$\}.
\end{algorithmic}
\end{algorithm}

\subsection{AdapEdit Algorithm Framework}
\begin{figure}[t]
	\centering
	\includegraphics[width=1.0\linewidth]{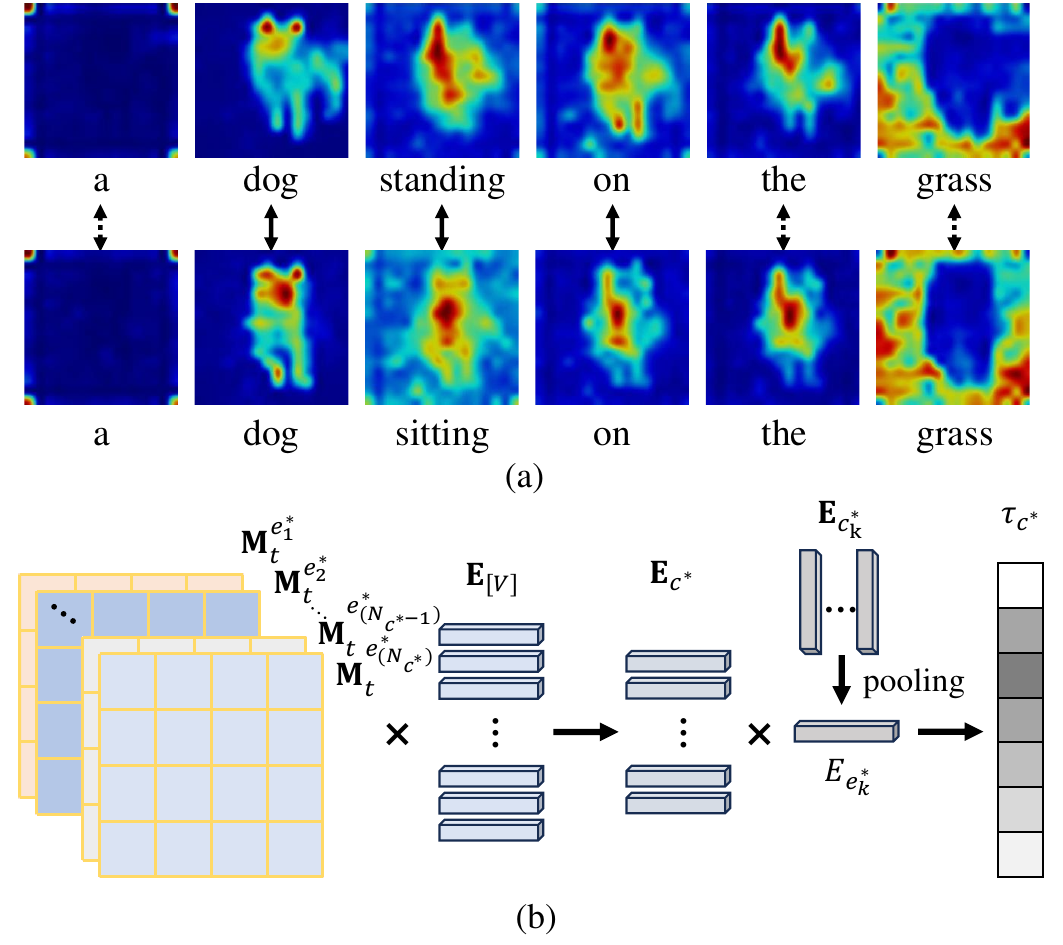}
	\caption{The illustration of our proposed soft attention strategy, in which (a) shows the cross-attention maps from $\bm{c}$ and $\bm{c}^*$ and 
            %
            %
            (b) details the specific calculating process.
            %
            }
	\label{fig: FWT}
\end{figure}

The pipeline of the proposed AdapEdit is illustrated in Figure~\ref{fig: model}.
To facilitate user-friendly editing, the model's input consists of two textual conditions $\{\bm{c}, \bm{c}^*\}$, which is the same with prompt-to-prompt~\cite{hertz2022prompt}, where $\bm{c}$ is the original text condition, and $\bm{c}^*$ denotes the edited condition.
Given the image $\textbf{x}$ that guidance-generated by $\bm{c}$, AdapEdit aims to manipulate hidden attention weights to modify the image $\textbf{x}$ to $\textbf{x}^*$ with guidance condition $\bm{c}^*$.
Most previous methods mainly focus \emph{hard editing} instructions such as adding, removing, or replacing the visual contents while basically ignore the more challenging \emph{soft editing} instructions (i.e., continuity-sensitive prompts).
To the end, we propose AdapEdit, a spatio-temporal guided adaptive editing algorithm.
Specifically, we first propose a flexible word-level temporal (FWT) adjustment module to support different guidance scales being adaptively assigned to sequential words for temporal-guided editing, which is driven by the impacts of each word $\bm{e}^*$ in the editing instruction $\bm{c^*}$ is variational. 
For further explanation, take the Figure~\ref{fig: FWT}(a) as an example, when only editing the attention map from $\textbf{M}_t^{\texttt{[standing]}}$ into $\textbf{M}_t^{\texttt{[sitting]}}$, the attention map from ``dog'' still remains the salient features of ``standing'', which hampers from the effective editing of the above instruction ``a dog sitting...''.

To address this issue, we design a soft attention strategy to calculate the $\tau_{\bm{c}^*}$ for fine-grained word-level editing in FWT module.
Further, we subsequently design a dynamic pixel-level spatial (DPS) weighting module to adaptively integrate the edited visual features into original image for spatial-guided editing. Note the spatial-guidance scales $\textbf{A}_\texttt{[S]}$ are determined by calculating the semantic similarities between the replaced words' average representations and the visual features.  
%
%
%
Based on the above two modules, AdapEdit can reliably control the DPMs (e.g., SD-v1.4) to effectively perform continuity-sensitive soft editing tasks with as many details as possible preserved from the original image $\textbf{x}$. Next, we detail the separate modules.

\begin{figure*}[t]
	\centering
	\includegraphics[width=1.0\linewidth]{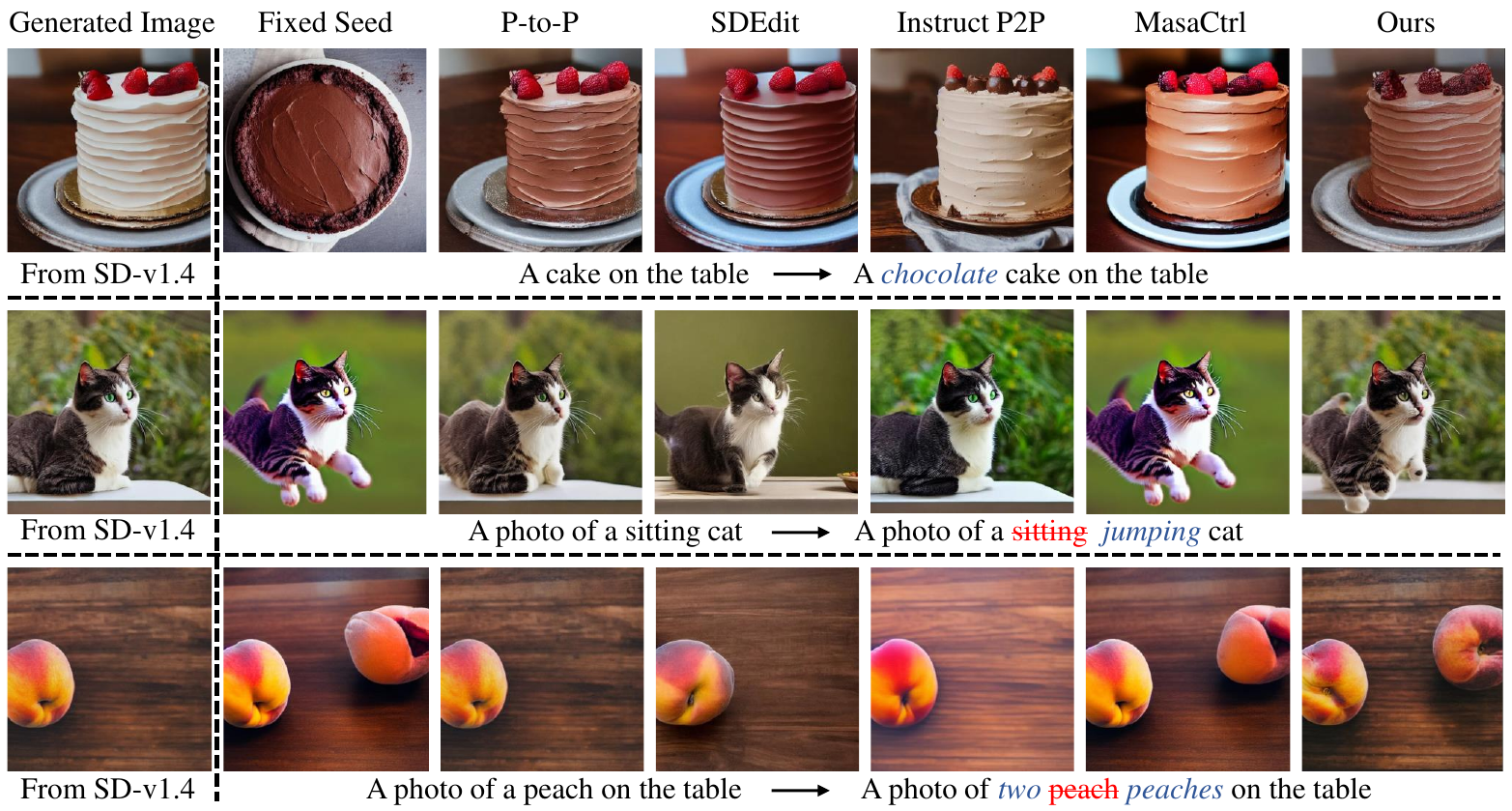}
	\caption{The qualitative comparisons of with the previous SOTA methods. The generated image (the leftmost column) denotes the original image $\textbf{x}$ conditioned on $\bm{c}$ generated by SD-v1.4, other columns present the editing results conditioned on $\bm{c}^*$. Note the fixed seed denotes generating a new image conditioned on $\bm{c}^*$ by directively using SD-v1.4 with the same random seed.
            %
            %
            %
            %
            }
	\label{fig: comparison}
\end{figure*}

\subsubsection{Flexible Word-Level Temporal Adjustment}
Flexible word-level temporal (FWT) adjustment module is adopted to adaptively calculate the temporal-guidance scales $\tau_{\bm{c}^*}$ for achieving temporal-guided editing. Instead using identical time truncation $\tau$ to diffusion models for the key words (here refer to the words that have been modified in $\bm{c}^*$ compared to $\bm{c}$) and general words, we first propose a soft attention strategy, depicted in Figure~\ref{fig: FWT} (b), to flexibly calculate $\tau_{\bm{e}^*_i}$ for each $\{\bm{e}^*_i\}_{i=1}^{N_{c^*}}$ ($N_{c^*}$ denotes the number of words in $\bm{c}^*$) in condition $\bm{c}^*$ to obtain updated attention map $\overline{\textbf{M}}_t^{\bm{e}_i^*}$.
Formally,
\begin{equation}
{\overline{\textbf{M}}_t^{\bm{e}_i^*}}=\left\{
\begin{aligned}
& \textbf{M}_t^{\bm{e}_{i}} &  t > \tau_{\bm{e}_i^*},  \\
& \bm{\mathcal{C}}({\textbf{M}_t^{\bm{e}_{i}}} , {\textbf{M}_t^{\bm{e}^*_{i}}}) & \text{otherwise}.
\end{aligned}
\right.
\label{eq5}
\end{equation}
where $\bm{e}_{i}$ and $\bm{e}_i^*$ respectively denotes $i^\text{th}$ word in $\bm{c}$ and $\bm{c}^*$, $\textbf{M}_t$ are their corresponding attention maps. When $t$ $\ge$ $\tau_{\bm{e}^*_i}$, the updated attention map $\overline{\textbf{M}}_t^{\bm{e}_i^*}$ equals to $\textbf{M}_t^{\bm{e}_{i}}$ for feature preservation of image $\textbf{x}$. 
While when $t$ $\textless$ $\tau_{\bm{e}^*_i}$, the updated map $\overline{\textbf{M}}_t^{\bm{e}_i^*}$ is obtained by the combination of $\textbf{M}_t^{\bm{e}_{i}}$ and $\textbf{M}_t^{\bm{e}^*_{i}}$, where $\bm{\mathcal{C}}(\cdot)$ represents the interpolation operation described in Eq.~\ref{eq9}, which will be detailed in the next DPS module.

Then, when running to the last diffusion step, we calculate the average of the cross-attention maps among all the layers based on 32$\times$32 resolution. 
After that, when given the cross-attention map $\{\textbf{M}_t^{\bm{e}^*_{i}}\}^{N_{\bm{c}^*}}_{i=1}$ of each word and visual embeddings $\textbf{E}_\texttt{[V]} \in \mathcal{R}^{N_\texttt{[V]} \times d}$,
each $\textbf{M}_t^{\bm{e}^*_{i}} \in \mathcal{R}^{1 \times N_\texttt{[V]}}$, $N_\texttt{[V]}$ is the number of visual pixels.
Followed by the softmax normalization layer, the $\{\textbf{M}_t^{\bm{e}^*_{i}}\}^{N_{\bm{c}^*}}_{i=1}$ are multiplied with the visual embeddings $\textbf{E}_\texttt{[V]}$ to obtain the textual embeddings $\textbf{E}_{\bm{c}^*} \in \mathcal{R}^{N_{\bm{c}^*} \times d}$ corresponding to the edited text condition $\bm{c}^*$ as,
\begin{equation}
\textbf{E}_{\bm{c}^*} = \textbf{M}_t^{\bm{c}^*} \times \textbf{E}_\texttt{[V]},
\label{eq6}
\end{equation}
Next, the embeddings of key words are fused by average pooling, which is defined as $\textbf{E}_{\bm{e}^*_k}$.
%
Then, after normalizing the embeddings, the correlation vector $\bm{A}_{\bm{c}^*}$ between the words $\{\bm{e}^*_i\}_{i=1}^{N_{c^*}}$ and the embeddings of replaced words $\textbf{E}_{\bm{e}^*_k}$ can be calculated by:
\begin{equation}
\bm{A}_{\bm{c}^*} = \textbf{E}_{\bm{c}^*} \times \textbf{E}_{\bm{e}^*_k}, \bm{A}_{\bm{c}^*} \in R^{1 \times {N_{\bm{c}^*}}},
\label{eq7}
\end{equation}
Finally, the calculation of $\tau_{\bm{e}^*_i}$ can be formally defined as:
\begin{equation}
\tau_{\bm{e}^*_i}=\left\{
\begin{aligned}
& 0, & \bm{e}^*_i \in \bm{c}^*_k, \\
& \lambda_{\tau_{\bm{c}^*}} [1 - exp(\bm{A}_{\bm{e}^*_i} - 1)], & \text{otherwise}.
\end{aligned}
\right.
\label{eq8}
\end{equation}
%

%
%
To conveniently control the range of $\tau_{\bm{c}^*}$, we leverage a hyper-parameter $\lambda_{\tau_{\bm{c}^*}}$ to adjust each $\tau_{\bm{e}_i^*}\in \tau_{\bm{c}^*}$.
In practice, we observe the $\tau_{\bm{e}^*_i}$ calculated by the above process is somewhat far from the expected result since the impacts of the background regions.
Taking Figure~\ref{fig: FWT} (a) as an example, the background region of ``sitting'' has large amount of overlaps with the foreground region of ``grass''.
The large amount of non-zero pixels in background region would interfere the $\bm{A}_{\texttt{[grass]}}$.
%
%
Therefore, a simple but effective strategy is utilized to obtain a mask for each $\textbf{M}_t^{\bm{e}^*_{i}}$.
When the values in $\textbf{M}_t^{\bm{e}^*_{i}} \textless \alpha_{m}$, we set them as 0.
The $\alpha_{m}$ is empirically set as $0.03$.
Eventually, we can obtain the discriminant embeddings to calculate the temporal-guidance scales $\tau_{\bm{c}^*}$.

\subsubsection{Dynamic Pixel-Level Spatial Weighting}
\label{DPS}
Dynamic pixel-level spatial (DPS) weighting module is adopted to adaptively calculate the spatial-guidance scales $\textbf{S}_\texttt{[V]}$ corresponding to visual content $\texttt{[V]}$ for achieving spatial-guided editing, here $\textbf{S}_\texttt{[V]}$ denotes the similarity matrix between the visual pixels of $\texttt{[V]}$ and editing tokens of $\bm{c}^*$.
%
%
Note since all the words $\bm{e}_i^* \in \bm{c}^*$ share the same $\textbf{S}_\texttt{[V]}$, 
the subscript $i$ will be omitted and the attention maps will be denoted as $\textbf{M}_{t}^{\bm{c}}$ and $\textbf{M}_{t}^{\bm{c}^*}$ as a whole if no confusion occurs in the following discussion. To adaptively integrate the original characteristics, inspired by ~\cite{tumanyan2023plug}, we introduce a hyper-parameter $\lambda_{S}$ to weight $\textbf{M}_t^{\bm{c}}$ and $\textbf{M}_t^{\bm{c}^*}$ and subsequently utilize an interpolation operation to integrate them as: 
\begin{equation}
\begin{aligned}
\bm{\mathcal{C}}({\textbf{M}_t^{\bm{c}}} , {\textbf{M}_t^{\bm{c}^*}}) = 
\lambda_{S} [{\textbf{S}_\texttt{[V]}} \cdot \textbf{M}_t^{\bm{c}^*} + {(1-\textbf{S}_\texttt{[V]}}) 
\cdot \textbf{M}_t^{\bm{c}}] \\ 
 + (1-\lambda_{S}) \textbf{M}_t^{\bm{c}}.
\label{eq9}
\end{aligned}
\end{equation}
where after normalizing the $\textbf{E}_{\bm{e}^*_k}$ (filtered by $\alpha_m$) and $\textbf{E}_\texttt{[V]}$, the similarity matrix $\textbf{S}_\texttt{[V]}$ is calculated as:
\begin{equation}
\textbf{S}_\texttt{[V]} = \lambda_{\textbf{S}_\texttt{[V]}} \textbf{E}_{\texttt{e}^*_k} \times \textbf{E}_\texttt{[V]}.
\label{eq10}
\end{equation}
Similar to $\lambda_{\tau_{\bm{c}^*}}$, we utilize $\lambda_{\textbf{S}_\texttt{[V]}}$ to adaptively adjust the range of values in $\textbf{S}_\texttt{[V]}$.
%
%
In sum, based on two modules FWT and DPS, AdapEdit can perform continuity-sensitive editings with the spatio-temporal guidance scales $\tau_{\bm{c}^*}$ and $\textbf{S}_\texttt{[V]}$ in a training-free manner.

\section{Experiments}
\subsection{Experimental Setup}
\subsubsection{Implementation Details}
We adopt the Stable Diffusion v1.4 version (SD-v1.4)~\cite{rombach2022high} as backbone for all experiments. Moreover, we follow~\cite{song2020denoising} to set the default denoising iterations as $50$. The classifier-free guidance scale is set to 7.5 and the maximum length of text instructions is set to 77. All experiments are conducted on 2 NVIDIA RTX3090 GPUs with PyTorch.

\begin{table}[t!]
    \footnotesize
    \centering
    \setlength{\leftskip}{-2pt}{
    \setlength{\tabcolsep}{4pt}
    \renewcommand\arraystretch{1.2}
    {
    \begin{tabular}{l ccc}
        \toprule
         & LPIPS $\downarrow$ & CLIP score $\uparrow$ & CLIP directional $\uparrow$ \\
         \hline
        P-to-P~(\citeyear{hertz2022prompt}) & 0.33 & 26.05 & 0.11 \\
        SDEdit (\citeyear{meng2021sdedit}) & 0.55 & 26.52 & 0.11  \\
        Instruct P2P~(\citeyear{brooks2023instructpix2pix}) & \textbf{0.27} & 24.59 & 0.04 \\
        MasaCtrl (\citeyear{cao2023masactrl}) & 0.41 & 27.00 & \textbf{0.13} \\
        AdapEdit (Ours) & 0.29 & \textbf{27.08} &  \textbf{0.13} \\
        \bottomrule
    \end{tabular}
    }
    }
    \caption{
    The quantitative evaluation results. 
    %
    }
    \label{tab:quantitative results}
\end{table}

\begin{figure}[t]
	\centering
	\includegraphics[width=1\linewidth]{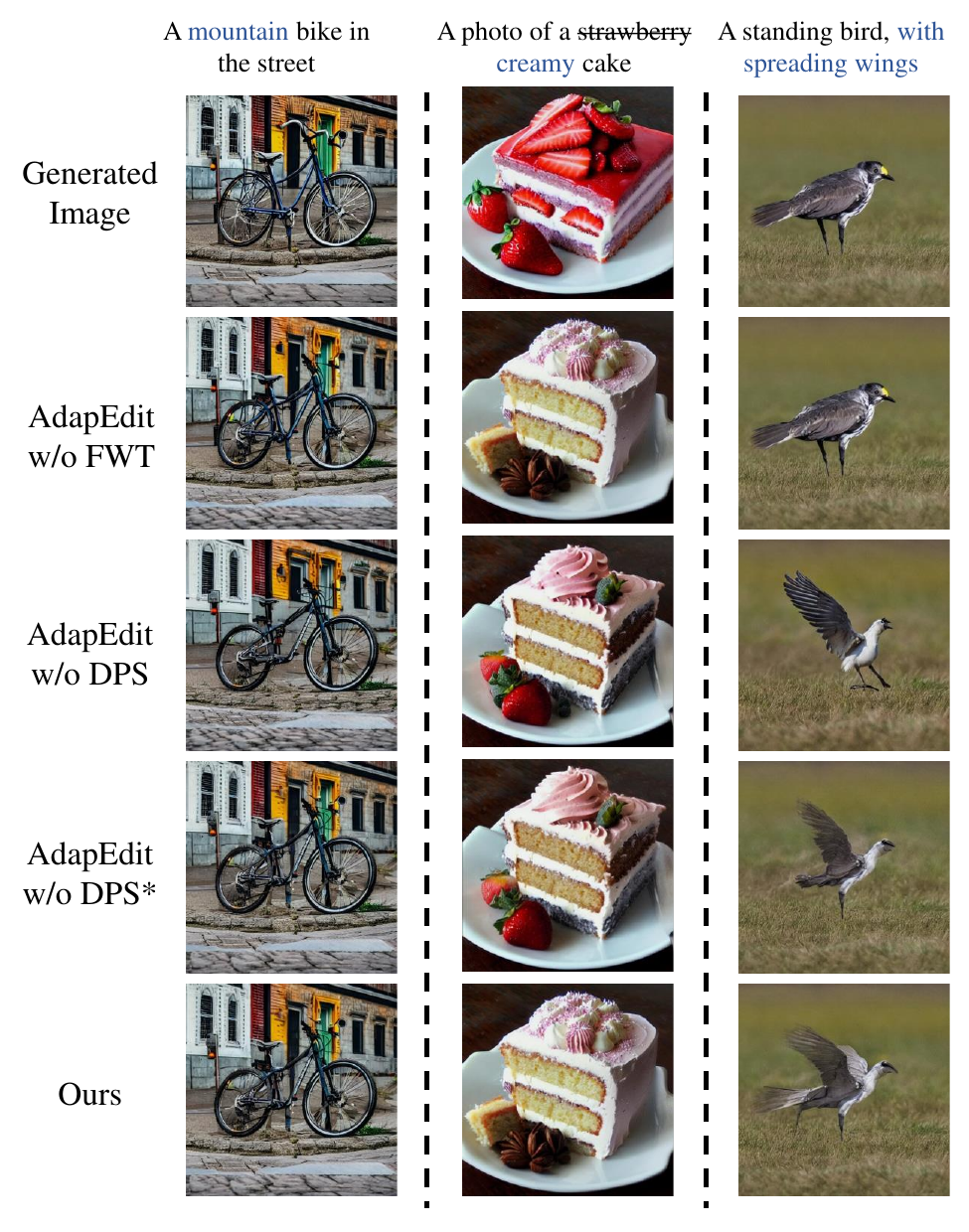}
	\caption{
    The ablation study of AdapEdit.
    }
	\label{fig: ablation}
\end{figure}

\begin{figure*}[t]
	\centering
	\includegraphics[width=1\linewidth]{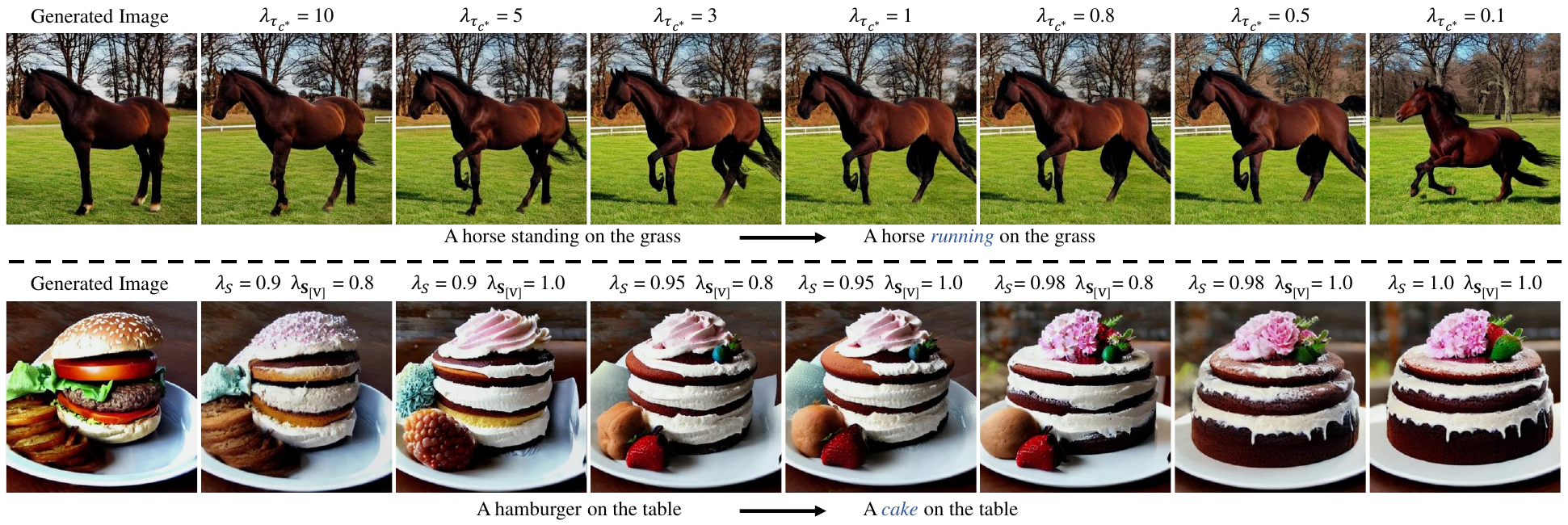}
	\caption{
    The hyper-parameter analysis of our proposed AdapEdit.
    We additionally introduce $\lambda_{\tau_{\bm{c}^*}}$ and $\lambda_{\textbf{S}_\texttt{[V]}}$ to control the range of $\tau_{\bm{c}^*}$ and $\textbf{S}_\texttt{[V]}$, and presents the results with different $\lambda_{S}$.
    }
	\label{fig: hyper-parameter}
\end{figure*}

\subsubsection{Baselines}
We compare AdapEdit versus the two groups of SOTA image editing methods. 1) The first group is real-image manipulation methods, which includes SDEdit~\cite{meng2021sdedit} and InstructPix2Pix~\cite{brooks2023instructpix2pix}. In the two baselines, the target text-condition $\bm{c}^*$ and the original real image $\textbf{x}$ server as common inputs. To be consistent with them, we utilize prompt-to-prompt~\cite{hertz2022prompt} to generate the original image for comparison. 2) The second group is text-generated methods, which includes Prompt-to-Prompt~\cite{hertz2022prompt} and MasaCtrl~\cite{cao2023masactrl}. Both of them use the original text-condition $\bm{c}$ and the target text-condition $\bm{c}^*$ as inputs, to generate the original image $\textbf{x}$ and edited image $\textbf{x}^*$, we keep the same settings. Note we have fixed the same random seed in our experiments for obtaining the same original images.

\subsection{Qualitative Evaluation}
%
%
%
The qualitative evaluation results are shown in Figure~\ref{fig: comparison}. From Figure~\ref{fig: comparison}, we obtain the following three observations.
\textbf{1)} For the hard editing instruction from \emph{``cake''} to \emph{``chocolate cake''}, all the methods show excellent performance, which is relatively easier to be achieved for diffusion-based methods. 
\textbf{2)} Further, when editing the posture of ``cat'' in the second line of Figure~\ref{fig: comparison} and the number of ``peach'' in the third line of Figure~\ref{fig: comparison}, Prompt-to-Prompt, SDEdit, and InstructP2P all fail to accomplish the posture- or number-oriented editing task. Benefiting from the adaptive editing capability of the FWT and DPS modules, our approach effectively copes with all hard or soft editing scenes. Moreover, as shown in Figure~\ref{fig: motivation}, AdapEdit can also perform adaptive editing to control that the \emph{``two apples''} have different sizes with certain reasoning capability emerging.
\textbf{3)} MasaCtrl achieves excellent object editing performance but fails to retain more details from the original image. It is worth noting that both of these are well addressed by our AdapEdit, which proves its effectiveness.

\subsection{Quantitative Evaluation}
Following~\citet{wang2023instructedit}, we further evaluate our approach in LPIPS, CLIP score and CLIP directional metrics.
Specifically, we use the dataset from InstructPix2Pix~\cite{brooks2023instructpix2pix}, which contains 700 human annotated samples and 454,445 editing instructions generated by GPT-3. To avoid the potential noises exist in the machine-generated instructions, we random select 700 samples annotated by human for evaluation.
From Table~\ref{tab:quantitative results}, we can find the AdapEdit achieves the highest CLIP score and CLIP directional similarity, while its LPIPS is only slightly higher than the InstructPix2Pix.
Unlike other methods, such as MasaCtrl, which performs well in CLIP score and CLIP directional similarity, but lacks consistency in editing results.
InstructPix2Pix has a low LPIPS value but performs poorly in CLIP score.
The above phenomenon demonstrates that our method strikes a good balance between the continuity and consistency of editing.
%
%
%

%

%

\subsection{Ablation Study}
In this part, we perform ablation experiments to evaluate the effectiveness of each component. We focus on four crucial settings: 1) w/o FWT denotes that we remove the flexible word-level temporal (FWT) adjustment module; 2) w/o DPS denotes that we remove the dynamic pixel-level spatial (DPS) weighting module; 3) w/o DPS$^*$ denotes that we only remove the above-mentioned pixel-level masking strategy in DPS. The visualized cases under given different settings are shown in the Figure~\ref{fig: ablation}.
From Figure~\ref{fig: ablation}, we can observe that removing each component will result in a performance degradation. Particularly, when removing DPS or DPS* module in column 2, our AdapEdit will be difficult to remove the \emph{``strawberries''} on the plate, which verfies the effectiveness of the DPS module. Further, when remove the pixel-level masking strategy in DPS (i.e., w/o DPS*), the algorithm will face a slightly inconsistency issue with unexpectedly changes to the original characteristics, which shows the proposed pixel-level masking strategy can effectively perceive the crucial regions corresponding to the target editing instruction and weaken the impacts of irrelevant regions. Moreover, when removing the FWT module in column 3, the bird standing on the grass will be difficult to be edit to \emph{``spread wings''}, which proves the effectiveness of our proposed FWT module in facing continuity-sensitive editing tasks.

\subsection{Hyper-parameter Analysis}
To analyze the impacts of temporal scales $\tau_{\bm{c}^*}$ and spatial scales $\textbf{S}_\texttt{[V]}$,we conduct the hyper-parameter experiments on $\lambda_{\tau_{\bm{c}^*}}$, $\lambda_{\textbf{S}_\texttt{[V]}}$, and $\lambda_{S}$. Note $\lambda_{S}$ is spatial interpolation weight. 
%
%
The exprimental results are presented in the Figure~\ref{fig: hyper-parameter}. As can be observed in the top row of Figure~\ref{fig: hyper-parameter}, the temporal hyper-parameter $\lambda_{\tau_{\bm{c}^*}}$ reveals the temporal continuity-editing process. Specifically, with the $\lambda_{\tau_{\bm{c}^*}}$ gradually decreases, the posture and movement of the horse are gradually becoming semantically close to the \emph{``running horse''}. Further, as illustrated by second and third edited images in the bottom row of Figure~\ref{fig: hyper-parameter}, we can observe that when keeping $\lambda_{S}=$ 0.9 unchanged and only vary $\lambda_{\textbf{S}_\texttt{[V]}}$ from 0.8 to 1.0, the \emph{``hamburger''} on the table shows significant variations with the cream on the top of \emph{``cake''} has emerged. Moreover, when the $\lambda_{S}$ reaches 0.98 and $\lambda_{\textbf{S}_\texttt{[V]}}$ reaches 1.0, the \emph{``hamburger''} shown in the picture has been significantly edited to \emph{``cake''}, which verifies the effectiveness of adaptive spatial editing and indicates it is necessary to choose appropriate hyper-parameters for controllable spatio-temporal guidance.
%
%

\section{Conclusion}
In this work, we first propose to focus on continuity-sensitive ``soft editing'' tasks and based on the task, we further propose a spatio-temporal guided adaptive editing algorithm, AdapEdit for short. Specifically, the algorithm mainly includes two elaborately designed modules, flexible word-level temporal (FWT) adjustment module and dynamic pixel-level spatial (DPS) weighting module, to achieve spatio-temporal guided adaptive editing. Based on the promising insight that adopting variable guidance degrees into forward diffusion process would facilitate fine-grained image editing at the pixel-level, we adopt a soft-attention strategy to achieve the challenging continuity-sensitive ``soft editing'' tasks. Moreover, we additional introduce a spatial interpolation weight for adaptively retaining the characteristics of the original images. Experiments on the real image editing dataset demonstrate the effectiveness and superior performance of our AdapEdit in both qualitative and quantitative evaluation.

\section*{Acknowledgments}
This work was supported by the National Key Research and Development Program of China (2022ZD0160603), the Project funded by China Postdoctoral Science Foundation (No. 2023M741950), and the Nationally Funded Postdoctoral Researcher Program (GZB20230347).

\bibliography{aaai24}

\end{document}